\pdfoutput=1
\documentclass[11pt]{article}

\usepackage[final]{acl}

\usepackage{times}
\usepackage{latexsym}

\usepackage[T1]{fontenc}
\usepackage[utf8]{inputenc}

\usepackage{microtype}

\usepackage{amsmath}
\DeclareMathOperator*{\argmin}{argmin} % thin space, limits underneath in displays
\usepackage{times}
\usepackage{graphicx}
\usepackage{color}
\usepackage{multirow}
\graphicspath{ {figures/} }

\title{Causal Graph in Language Model Rediscovers Cortical Hierarchy in Human Narrative Processing}

\author{Zhengqi He \textsuperscript{a} \thanks{\phantom{.}  zhengqi.he@riken.jp} 
        \and 
        Taro Toyoizumi \textsuperscript{a,b} \thanks{\phantom{.}  taro.toyoizumi@riken.jp} \\ 
        \\ 
        \textsuperscript{a} Lab for Neural Computation and Adaptation, RIKEN Center for Brain Science, Japan\\
        \textsuperscript{a,b} Department of Mathematical Informatics, Graduate School of Information Science \\
          and Technology, The University of Tokyo, Japan\\
           }

\begin{document}
\maketitle
\begin{abstract}

Understanding how humans process natural language has long been a vital research direction. The field of natural language processing (NLP) has recently experienced a surge in the development of powerful language models. These models have proven to be invaluable tools for studying another complex system known to process human language: the brain. Previous studies have demonstrated that the features of language models can be mapped to fMRI brain activity. This raises the question: is there a commonality between information processing in language models and the human brain? To estimate information flow patterns in a language model, we examined the causal relationships between different layers. Drawing inspiration from the workspace framework for consciousness, we hypothesized that features integrating more information would more accurately predict higher hierarchical brain activity. To validate this hypothesis, we classified language model features into two categories based on causal network measures: 'low in-degree' and 'high in-degree'. We subsequently compared the brain prediction accuracy maps for these two groups. Our results reveal that the difference in prediction accuracy follows a hierarchical pattern, consistent with the cortical hierarchy map revealed by activity time constants. This finding suggests a parallel between how language models and the human brain process linguistic information.

\end{abstract}

\section{Introduction}

Understanding high-order cognitive functions of the human brain, such as natural language processing, remains a pivotal challenge in neural science \citep{hickok2007cortical, friederici2011brain, ralph2017neural}. Modern neuroimaging techniques, like functional Magnetic Resonance Imaging (fMRI), allow us to observe brain activity during language-related tasks directly. A prevailing hypothesis in this domain is the hierarchical processing hypothesis. A seminal study supporting this hypothesis is presented in \citep{lerner2011topographic}. In this study, the author investigates the effects of scrambling language elements at different hierarchical levels—ranging from words to sentences to paragraphs. The findings reveal distributed networks of brain areas to accumulate language information over time, emphasizing the hierarchical nature of language processing.

It's noteworthy that the correlation between hierarchy and time constant appears to be a general characteristic, not exclusive to language \citep{huntenburg2018large,raut2020hierarchical}. For instance, in studies by \citep{hasson2008hierarchy, honey2012slow}, a hierarchical temporal receptive window was identified in humans while watching movies, as observed through fMRI and ECoG recordings. Similarly, \citep{murray2014hierarchy} uncovered a hierarchy of intrinsic time scales in the cortex of macaque monkeys, evidenced by spike train recordings. Collectively, this body of research suggests a linkage between temporal properties and ranks within the cortical hierarchy. It's hypothesized that brain regions with a slower time constant typically occupy higher ranks in anatomically defined hierarchy \citep{felleman1991distributed, barbas1997cortical, markov2014anatomy}.

Besides, Deep Neural Networks, which draw inspiration from the brain's computational principles, have achieved significant success in the domain of natural language processing. Recent trends highlight the rise of large unsupervised language models (LMs), such as ELMo \citep{peters-etal-2018-deep}, GPT \cite{radford2018improving}, and BERT \cite{devlin2018bert}. Subsequently, plenty of research has delved into harnessing their potential through various methods, including the pretraining-finetuning paradigm \cite{devlin2018bert}, prompt-engineering paradigm \citep{brown2020language}, and the development of chatbots \citep{ouyang2022training, openai2023gpt}.

Historically, language studies using brain imaging often relied on tightly controlled conditions that, while simple, may not always mimic natural scenarios and might not be easily generalizable \citep{lerner2011topographic}. A compelling question that arises is whether the computational mechanisms of deep neural networks and the brain can be compared. A pioneering study by Daniel Yamins \citep{yamins2014performance} showed that deep neural networks, even when not specifically trained to emulate neural activity, exhibited patterns highly predictive of brain activity in areas like the V4 and inferior temporal cortex when trained on object categorization tasks. This approach paved the way for comparing neural networks to brain activities during natural language processing. An early work is represented by \citep{huth2016natural}. The authors demonstrated that features derived from word embeddings could map onto cortical activity during natural speech processing. In another study, \cite{schrimpf2021neural} compared brain imaging data from individuals reading natural language materials to various language models, spanning from basic embeddings to complex neural networks. Interestingly, models with superior language prediction capabilities also tended to predict brain activity better.  With modern deep neural network-based language models, the complicated dynamics of natural language can now be encoded and compared directly with brain data, which introduces exciting avenues for novel discoveries.

Given that features in a language model (LM) can map to whole-brain activity, a natural question arises: is there a fundamental similarity in information processing between an LM and the brain, or is the correlation merely a superficial coincidence \citep{antonello2023predictive}? It's well-established that the middle to late layers of a multi-layer transformer-based LM often align best with brain activity across both low and high hierarchical regions. However, the information in these hierarchical brain areas possesses distinct properties. For instance, according to the workspace framework for consciousness \citep{dehaene2001towards}, higher cortical brain regions typically integrate information from a greater number of source areas compared to lower hierarchical regions. Inspired by this observation, we hypothesize that, if LM and the brain share similarity in information processing, part of the language features in the middle-late layers of LM that integrate from a more diverse range of source features, are more likely to predict activity in higher brain hierarchies, and vice versa. We sought to validate this hypothesis by approximating the information flow in an LM using a causal graph. We argue that features integrating from a broader array of source features will possess a higher in-degree in such a causal graph. By grouping features based on in-degree measurements and fitting brain activity separately, we aimed to ascertain if these feature groups corresponded with the cortical hierarchy, potentially inferred through activity time scales.

\section{Related Works}

Our research intersects with two principal areas of study.

\textbf{Hierarchy in brain}. A body of work highlights the notion of an increasing time constant or temporal receptive field as a core organizing principle for the brain. For instance, \citep{murray2014hierarchy} unveiled an ascending intrinsic time scale within the cortical hierarchy, observed through auto-correlation measurements in the primate cortex. Meanwhile, the study by \citep{chaudhuri2015large} developed a comprehensive dynamical model of the macaque neocortex using a connectome dataset, shedding light on intrinsic time scale hierarchies. In \citep{baldassano2017discovering}, the authors explored the alignment between event structures featured with increasing time windows and cortical hierarchy, using human narrative perception datasets. Complementing this, \citep{chang2022information} identified a hierarchy in processing timescales via response lag gradients that correlate with known cortical hierarchies.

\textbf{Language model fitting brain}: Another line of research underscores the potential of language models in predicting human brain activity. Building upon the findings of \citep{huth2016natural}, which established that static word embeddings correlate with brain activity, subsequent studies demonstrated that contextualized word representations surpassed their static counterparts in terms of accuracy in predicting brain activity, as indicated by \citep{jain2018incorporating}. There has since been an increasing trend of  studies comparing language models with brain activity datasets \citep{toneva2019interpreting, schrimpf2021neural, goldstein2022shared, kumar2022reconstructing, caucheteux2022brains, millet2022toward}. Concurrently, innovative strategies aimed at augmenting the alignment between language models and brain recordings have been proposed \citep{schwartz2019inducing, aw2022training, antonello2023scaling}. Comprehensive reviews and summaries of these studies are articulated in works such as \citep{abdou2022connecting, arana2023deep, jain2023computational}. Notably, the concept of hierarchy is recurrently discussed within this domain. For instance, \citep{jain2020interpretable} introduced a multi-timescale LSTM, capturing the temporal hierarchy observed in natural speech fMRI datasets, while \citep{caucheteux2023evidence} explored the relationship between cortical hierarchy and enhancements in brain activity predictions across varied predictive time windows.

%\textbf{Causality in the Brain}: The exploration of causality within brain activity is also a prominent topic of interest. Given the inherent challenges in directly tracing causality in the brain, researchers frequently employ proxy methodologies. For instance, \citep{londei2007brain} endeavored to construct a functional relationship graph for a passive word listening task, leveraging Granger-causality. Meanwhile, \citep{tajima2015untangling} intertwined the concepts of causality, directionality, and complexity, drawing upon the cross-embedding method as a defining metric.

\section{Methods}

\subsection{fMRI dataset}

We have selected the "Narratives" fMRI dataset as our primary dataset \citep{nastase2021narratives}. This dataset offers an extensive collection of fMRI recordings representing human brain activity as participants passively engage with naturalistic spoken narratives. It includes data from 345 participants who listened to a total of 27 distinct stories. In total, the dataset spans 1.3 million words, 370K repetition times (TRs), and 6.4 days of accumulated data across all participants. Each recording follows a consistent repetition time of 1.5 seconds. Preprocessing ensures that all fMRI data are smoothed, surfaced, and uniformly aligned to a shared space known as "fsaverage," which serves as the foundation for our subsequent analysis. Additionally, every story comes with a timestamped transcript, enabling us to process through language models, obtain contextualized word features, and synchronize them with the corresponding fMRI data.

\subsection{Mapping language features onto brain}
\label{sec:brain_fit}

In aligning language models with brain data, we adopted methodologies similar to those detailed in \citep{toneva2019interpreting, schrimpf2021neural, jain2020interpretable, caucheteux2023evidence}. Given that the "Narratives" dataset captures brain activity from multiple participants exposed to natural language stimuli, we introduce the same stimuli into a pre-trained language model and extract encoded representations from multiple layers. We mainly use the OPT-125m \citep{zhang2022opt}, a publicly available auto-regressive language model built on transformer architecture. The "Narratives" dataset has a resolution of 1.5 seconds per TR, while the features we extract from the language model are per token. To establish a meaningful comparison, we need to align the two datasets properly. Utilizing the timestamp of each token, we correlate it to a specific TR and then average the extracted features for a more comprehensive analysis.

Assuming we've reached this step, we have a time series of high-dimensional language model features represented as \(X_t\) with shape \((T, d)\). Here, \(T\) denotes the number of time steps, and \(d\) represents the number of dimensions for the language model features. This feature series has been aligned with our fMRI data \(W_t\) with shape \((T, l)\), where \(l\) stands for the number of voxels. Our subsequent task is to benchmark \(X_t\) and \(W_t\) using ridge regression. Given the high dimensionality of \(X_t\) (for instance, OPT-125m can have a dimension \(d\) as large as 768), we employ PCA to reduce the dimension of the representation vector for computational efficiency. In our implementation, we've reduced the dimension to 20, following \citep{caucheteux2023evidence}, balancing both prediction accuracy and computational speed.

We predict the activity of each fMRI voxel using a linear projection of the representation vector from different layers. This linear projection is regularized using ridge loss. The process of ridge regression is described as follows. Assume we have train and validation split of both language model features $X \rightarrow (X_\mu, X_\nu)$ and fMRI dataset $W \rightarrow (W_\mu, W_\nu)$, we first do ridge regression of the train split. Ridge regression can be described as a minimization problem for each voxel $i$:
\begin{equation}
\argmin_{V_i}(W_\mu^i-X_\mu V_i)^T(W_\mu^i-X_\mu V_i) + \alpha_i V_i^T V_i
\end{equation} 
where $\alpha_i$ is a regularization factor, $V_i$ is the fitting vector, $W_\mu^i$ is time series for voxel $i$. Then, the fitting vector is
\begin{equation}
V_i = (X_\mu^T X_\mu + \alpha_i I)^{-1} X_\mu^T W_\mu^i
\end{equation} 
Prediction accuracy is quantified by the correlation of the predicted brain signal with the measured brain signal on the validation split:
\begin{equation}
P(X, W) = \mathrm{Corr}(X_\nu V, W_\nu)
\end{equation} 
where $\mathrm{Corr}$ is the Pearson correlation operator. To use data efficiently, we perform multi-fold leave-one-out cross-validation, and the average accuracy among all folds is reported. The regularization factor is separately chosen from log-spaced between $10^{-1}$ and $10^8$ for each voxel via an extra nested leave-one-out cross-validation process. 

To account for the slow bold response, we also use the finite impulse response (FIR) model following \citep{huth2016natural} by concatenating language representation with delays from -9 to -3 TRs. The afni-smoothed version of the Narratives dataset is used in our study.

\subsection{Causal graph in language model features}
\label{sec:causality}

Our analysis is based on pretrained multi-layer, transformer-based, auto-regressive language models, such as OPT. The architectural design of these models facilitates the flow of information from early nodes to later ones and from the bottom layer to the top. Given that previous research has indicated that the middle-to-late layers of a language model align best with brain activity, our objective is to delve in detail into these findings. Specifically, we aim to find out which features of the model correspond to which parts of the brain. In this regard, we propose a causality measure. Using this measure, we can categorize language model features into 'low in-degree' and 'high in-degree' groups, which will be defined later, subsequently showcasing their relationship with brain hierarchy.

We use random noise perturbation to estimate causality. Consider a lower layer of interest, denoted by \( X = [x_1, x_2, ..., x_T] \), and a higher layer of interest, denoted by \( Y = [y_1, y_2, ..., y_T] \). Due to the inherent network structure of our language model, there's a general causal relationship such that \( X \rightarrow Y \), implying \( Y = f(X) \). Introducing a random perturbation $dX$ yields \( Y + dY = f(X + dX) \).

As mentioned earlier, both \( Y \) and \( X \) typically have high dimensions. Prior to fitting to the brain, we employ Principal Component Analysis (PCA) for dimensionality reduction. Following this approach, we further use PCA to reduce dimensions and then evaluate the causal relationship in the PCA-reduced space. Let's denote the transformed spaces as \( \bar{X} = PCA(X) = X M_x\) and \( \bar{Y} = PCA(Y) = Y M_y \). Utilizing the PCA projection matrices, perturbations and responses in the PCA space are determined as \( d\bar{X} = dX M_x\) and \( d\bar{Y} = dY M_y\) respectively. Subsequently, we obtain the causality matrix with time-shift $\tau$ as:
\begin{equation}
C_\tau = d\bar{Y}^{T}d\bar{X}_\tau/(T-\tau)
\end{equation} 
where $C_\tau$ is the causality matrix with time-shift $\tau$, $d\bar{X}_\tau$ represents $d\bar{X}$ with time-shift $\tau$. 

To construct a causal graph, we sum up the absolute value of the causality matrix for each $\tau$. Then we threshold the causality matrix by the median of the matrix elements; any value exceeding this threshold is considered a valid causal link. Finally, we obtain the causality matrix:
\begin{equation}
C = \mathrm{td}[\sum_\tau{\mathrm{abs}(C_\tau)}]
\end{equation} 
where $\mathrm{td}$ is the threshold operator, $\mathrm{abs}$ is the absolute operator.

\subsection{Mapping Causality onto Cortical Hierarchy}

\begin{figure}
    \centering
    \includegraphics[width=1\linewidth]{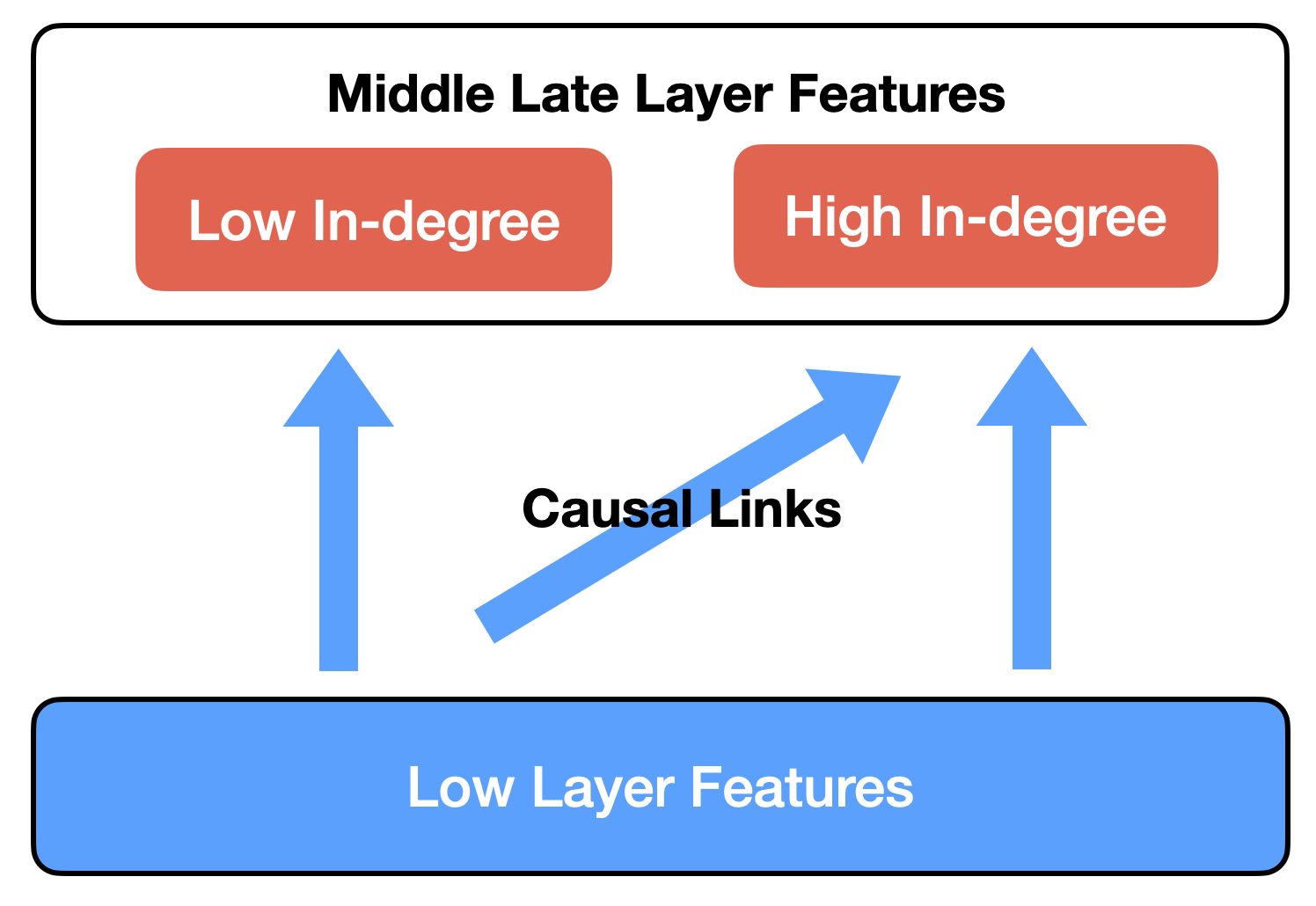}
    \caption{The causal relationships spanning across layers serve as a mechanism to differentiate features of varying in-degrees. Low in-degree features present in the middle late layer receive less causal influence compared to their high in-degree counterparts. We hypothesize that low in-degree features better predict low-hierarchical cortical areas, and vice versa.
    }
    \label{fig:causal_illustrate}
\end{figure}

In this section, we describe the rationale for using a causal graph to mirror the brain's hierarchy. As stated by the workspace framework of consciousness \citep{dehaene2001towards}, higher cortical areas, believed to host consciousness, integrate information from a broader array of source regions in the brain. Drawing from this perspective, we hypothesize that if a language model processes information analogously to the brain, its features would exhibit distinct patterns in terms of information integration. Specifically, features mirroring high cortical regions should integrate more information from preceding layers, and conversely for those resembling lower cortical areas. This conceptual framework is depicted in Fig. \ref{fig:causal_illustrate}. 

A feature group that integrates a greater variety of information from preceding layers would exhibit a higher in-degree of causal links, and conversely, those integrating a less variety of information would have fewer. Upon distinguishing between low in-degree feature groups and high in-degree feature groups, we can project each feature group onto cortical activity using the methodology described in Sec. \ref{sec:brain_fit}. Subsequent to this analysis, we can compare the resulting predicted brain maps to see if they align with the cortical hierarchy. This hierarchy is gauged using the activity time constant, a concept described in the following section.

%First, let's delve conceptually into partitioning language features using causality. The foundational concept behind this is depicted in Fig. \ref{fig:causal_illustrate}. We assume each layer as a mixture of different feature orders, including low-order features, which resides at source side of the causal link, and high-order features, which resides at sink side of the causal link. Although, topologically, lower layers consistently influence higher layers, our findings suggest that features in the lower layers don't exert a uniform influence on features in the upper layers. Consequently, our hypothesis is that, even if the network topology suggests that high-order features in lower layers can influence low-order features in upper layers, the magnitude of this causal effect remains relatively low. Hence, low-order features in upper layers are likely subjected to a weaker total causal impact compared to their high-order counterparts. Based on this idea, we've formulated a criterion to segregate low-order features from high-order ones. Thus, we define "low-order" features as features at high layers with "low indegree", and "high-order" features as features with "high indegree". We can similarly define features at low layers with "outdegree", which will be discussed in the appendix.

\subsection{Calculating fMRI time constant}
\label{sec:temporal}

As a reference baseline for hierarchy, we calculate the time constant for each brain voxel within the Narratives dataset. Our method aligns with the approach in \citep{murray2014hierarchy}, which leverages auto-correlation.

Given a time series for a particular voxel, \(W_t\), the auto-correlation at lag \(\tau\) is computed as:
\begin{equation}
\mathrm{AC}(W, \tau) = \mathrm{Corr}(W_{t}, W_{t-\tau})
\end{equation} 
Here, \(\mathrm{Corr}\) denotes the correlation coefficient. As \(\tau\) increases, we typically anticipate a decline in \(\mathrm{AC}\). Accordingly, we can model \(\mathrm{AC}(W, \tau)\) using an exponentially decreasing function:
\begin{equation}
\argmin_{\lambda}[ \mathrm{exp}(\tau/\lambda) - \mathrm{AC}(W, \tau) ]^2
\end{equation} 
Subsequently, the fitted coefficient \(\lambda\) serves as a representation of the voxel's time constant. By iterating over all voxels, we can generate a cortical map showcasing time constants across the Narratives dataset.

\section{Results}

\subsection{Language Model Features Predict Brain Activity}

\begin{figure}[t]
\includegraphics[width=1.0\linewidth]{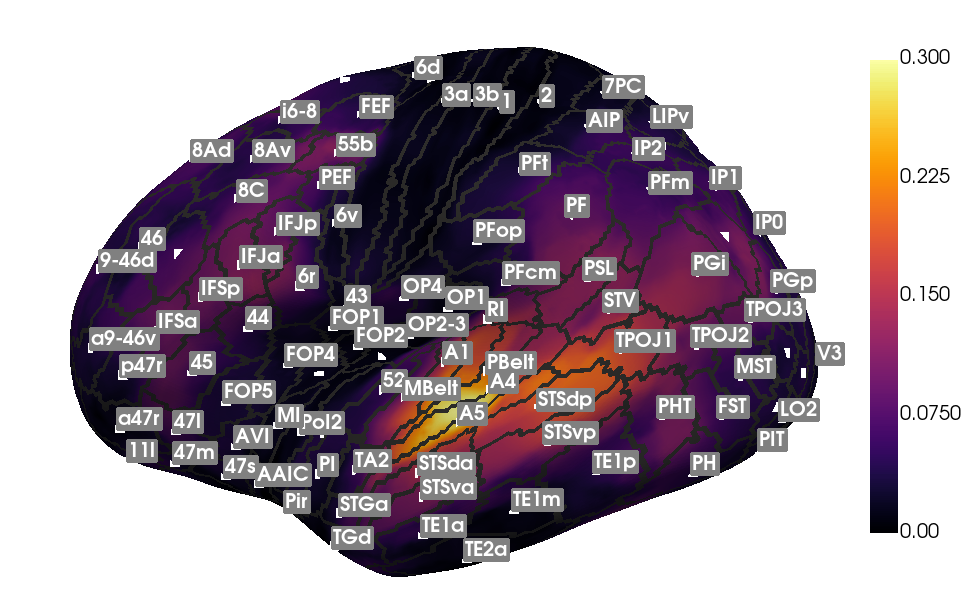}
\includegraphics[width=1.0\linewidth]{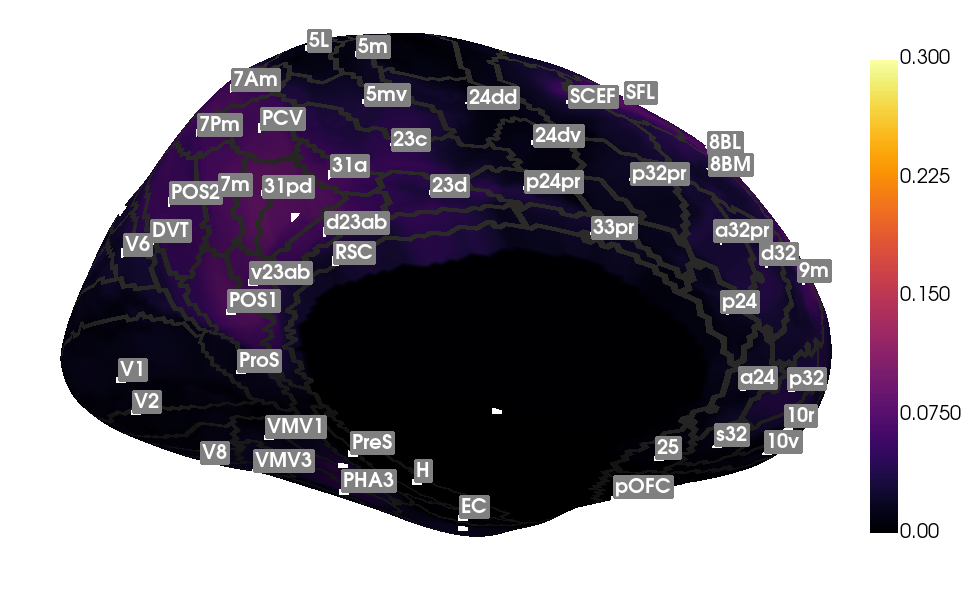}
\caption{Brain prediction accuracy map measured with correlation shown in fsaverage space, with boundaries and labels from Glasser atlas.}
\label{fig:brain_score}
\end{figure}

In our initial efforts, we sought to replicate the findings presented in prior literature, focusing on the predictive efficacy of language model features in relation to the fMRI dataset. The brain prediction accuracy, quantified using correlation coefficients, is illustrated in Fig. \ref{fig:brain_score}. This result captures the average across all participants, displayed on a unified "fsaverage6" surface template. To enhance visualization, boundaries and labels sourced from the Glasser Atlas \citep{glasser2016multi} have been incorporated. The 3D mesh visualization was achieved using the Python-based 3D-rendering engine, PyVista \citep{sullivan2019pyvista}.

It is evident from the plot that the prediction accuracy map is in line with findings from previous investigations. Notably, the strongest correlations emerge from lower auditory regions, specifically A4 and A5. This is closely followed by correlations within the language network, encompassing areas such as Broca's regions 44 and 45, along with segments of the temporal lobe like STSda, STSva, STSdp, and STSva. Subsequently, we observe notable correlations in high-order regions: within the frontal lobe areas like IFSa, IFSp, IFJa, and IFJp, and within the parietal lobe sectors such as PF, PFm, and PGi. The cumulative average correlation across all voxels is 0.0429 for layer 9, which corresponds to the late middle segment of the OPT-125m model comprising 12 layers.

To test the statistical significance of our findings, we adopted a shuffling approach for the language features and repeated the fitting process. This approach yielded a null result, characterized by approximately Gaussian distributed prediction accuracy with a mean of 0 and a standard deviation of 0.003. Given that the average value of the accuracy map is around 0.04—a magnitude ten times greater than the standard deviation of the null-hypothesis accuracy distribution—it becomes evident that the precision map derived through this method has statistical significance.

Furthermore, our exploration reemphasizes a previously observed trend specific to multi-layer transformer-based auto-regressive models like OPT: the predictive power of brain activity initially escalates with layer progression until middle-late layers. For instance, when the OPT 125m model fits the Narratives dataset, the average prediction accuracy manifests as 0.0268, 0.0352, 0.0429, and 0.0403 for layers 1, 5, 9, and 12, respectively. Given that layer 9 exhibits the peak, our subsequent analyses are based on this layer.

\subsection{Causal Graph Reveals Cortical Hierarchy}
\label{sec:causal_res}

\begin{figure}
    \centering
    \includegraphics[width=1\linewidth]{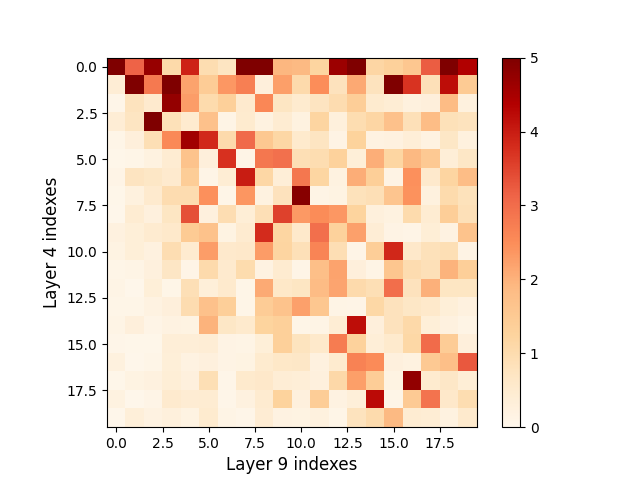}
    \caption{Causality matrix $C$. An entry at row \(i\) and column \(j\) quantifies the causal influence of dimension \(i\) in layer 4 on dimension \(j\) in layer 9.}
    \label{fig:causal_mat}
\end{figure}

We employed the methodology delineated in Sec. \ref{sec:causality} to find out the causal relationships among pairs of layers within the language model. Fig. \ref{fig:causal_mat} presents the derived causality matrix \(C\) from layer 4 to layer 9 of the Opt-125m model. The chosen number of dimensions for PCA is 20. In the matrix, an entry at row \(i\) and column \(j\) quantifies the influence of dimension \(i\) in layer 4 on dimension \(j\) in layer 9. To construct a causal graph, we applied a thresholding technique. Specifically, if \(C_{ij}\) surpasses the threshold defined as the median value of the causality matrix, we identify dimension \(i\) of layer 4 as posing a significant causal impact on dimension \(j\) in layer 9. By summing across dimension \(i\), we derive a vector that represents the number of causal links for each dimension \(j\) of layer 9. Those dimensions in layer 9 with fewer inbound causal links are categorized as "low in-degree" dimensions. Conversely, dimensions with a higher count of inbound causal links are classified as "high in-degree" dimensions. We designate the dimensions within the lower half as low in-degree, and those within the upper half as high in-degree features.

\begin{figure}[t]
\includegraphics[width=1.0\linewidth]{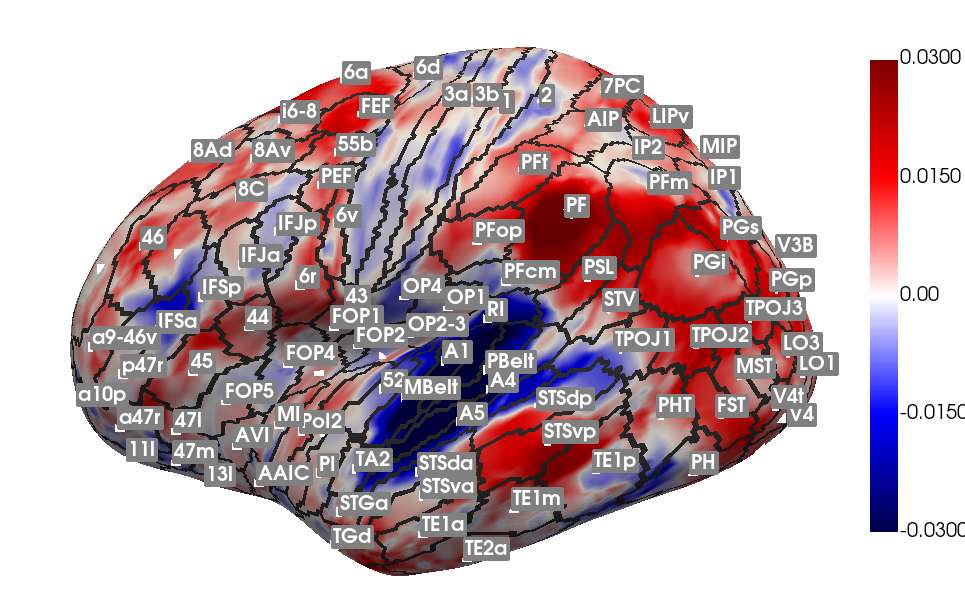}
\includegraphics[width=1.0\linewidth]{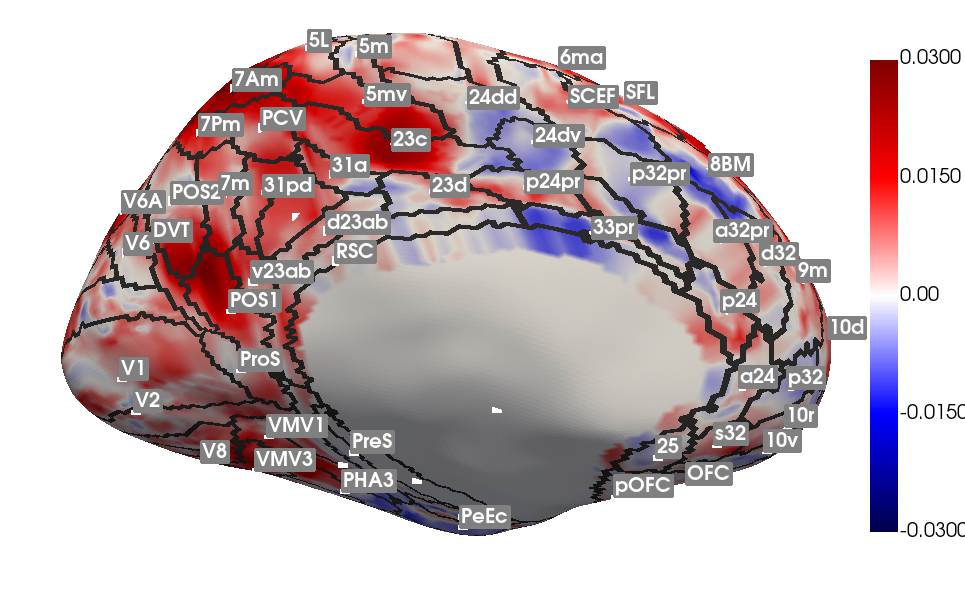}
\caption{Brain prediction accuracy difference between high in-degree feature group and low in-degree feature group measured with correlation, layer 9.}
\label{fig:causal_diff}
\end{figure}

Having partitioned the language model features into low in-degree and high in-degree categories, we then proceeded to predict brain activity for each category separately, utilizing the methodology previously discussed in Sec. \ref{sec:brain_fit}. This process yielded two distinct prediction accuracy maps, akin to the one illustrated in Fig. \ref{fig:brain_score}. To emphasize the distinct regional preferences, we computed the difference between these two maps, specifically by subtracting the low in-degree map from the high in-degree map. The resultant map is presented in Fig. \ref{fig:causal_diff}.

From the figure, it is evident that the accuracy maps produced by the high in-degree and low in-degree feature groups display notable differences. The color-coding, based on the subtraction of low in-degree from high in-degree features, reveals that regions colored in red are better predicted by the high in-degree group, while those in the opposite spectrum are predicted by the low in-degree group. Specifically, lower hierarchical regions proposed in \citep{lerner2011topographic} such as A4, A5, STSdp, 44, and 45 tend to align more closely with the low in-degree feature group. In contrast, higher hierarchical regions like PF, PFt, 6r, and 7m, are better represented by the high in-degree feature group.

To assess the statistical significance of our observations, we implemented a text-random shuffling technique. We predicted brain activity based on the shuffled language model features along text direction, and computed their brain prediction accuracy maps. Their differential accuracy map, obtained by subtracting two maps, follows a Gaussian distribution with a mean of 0 and a standard deviation of 0.004. Given that this deviation is much smaller than typical values observed in Fig. \ref{fig:causal_diff}, our findings can be considered statistically robust.

Furthermore, the robustness of our method across various layers and models is demonstrated in Appendix Sec. \ref{asec:robustness}.

\subsection{Time Constant Reveals Temporal Hierarchy}

While our initial hierarchy assessment was predicated on the language model fitting brain activity, a question emerges: Can we directly correlate this LM in-degree mapping to the hierarchy structure of the narrative brain data via the activity time constant?

The hierarchy we talked about in the previous section proposed in \citep{lerner2011topographic} is calculated based on narratives scrambled at different time scales, which cannot be reproduced in our case. Contemporary literature largely supports the notion that hierarchy correlates with activity time constants \citep{raut2020hierarchical}. Following this, we show that the hierarchy deduced from causality in Sec. \ref{sec:causal_res} reproduces the hierarchy inferred from activity time constants directly derived from the Narratives fMRI dataset.

\begin{figure}[t]
\includegraphics[width=1.0\linewidth]{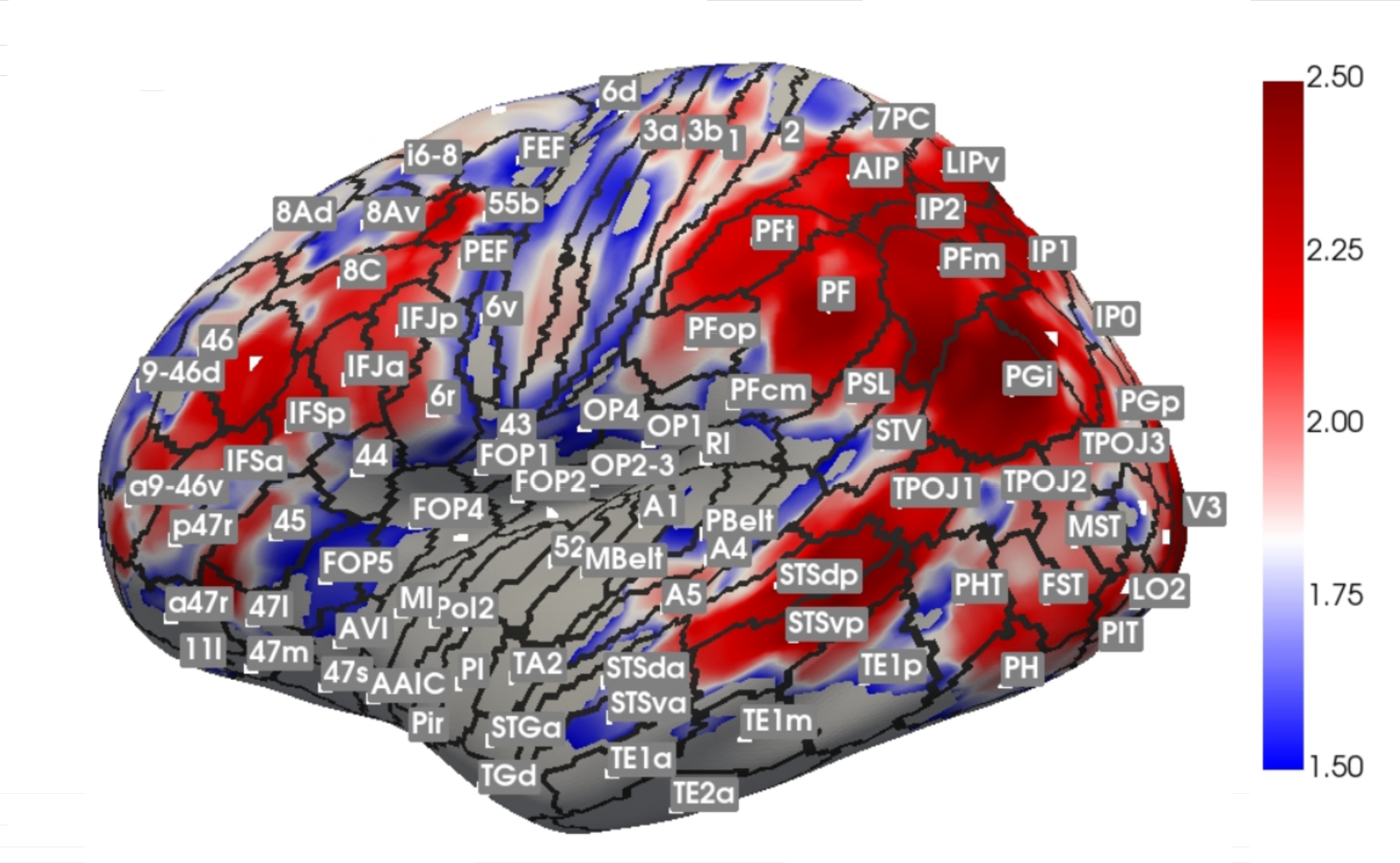}
\includegraphics[width=1.0\linewidth]{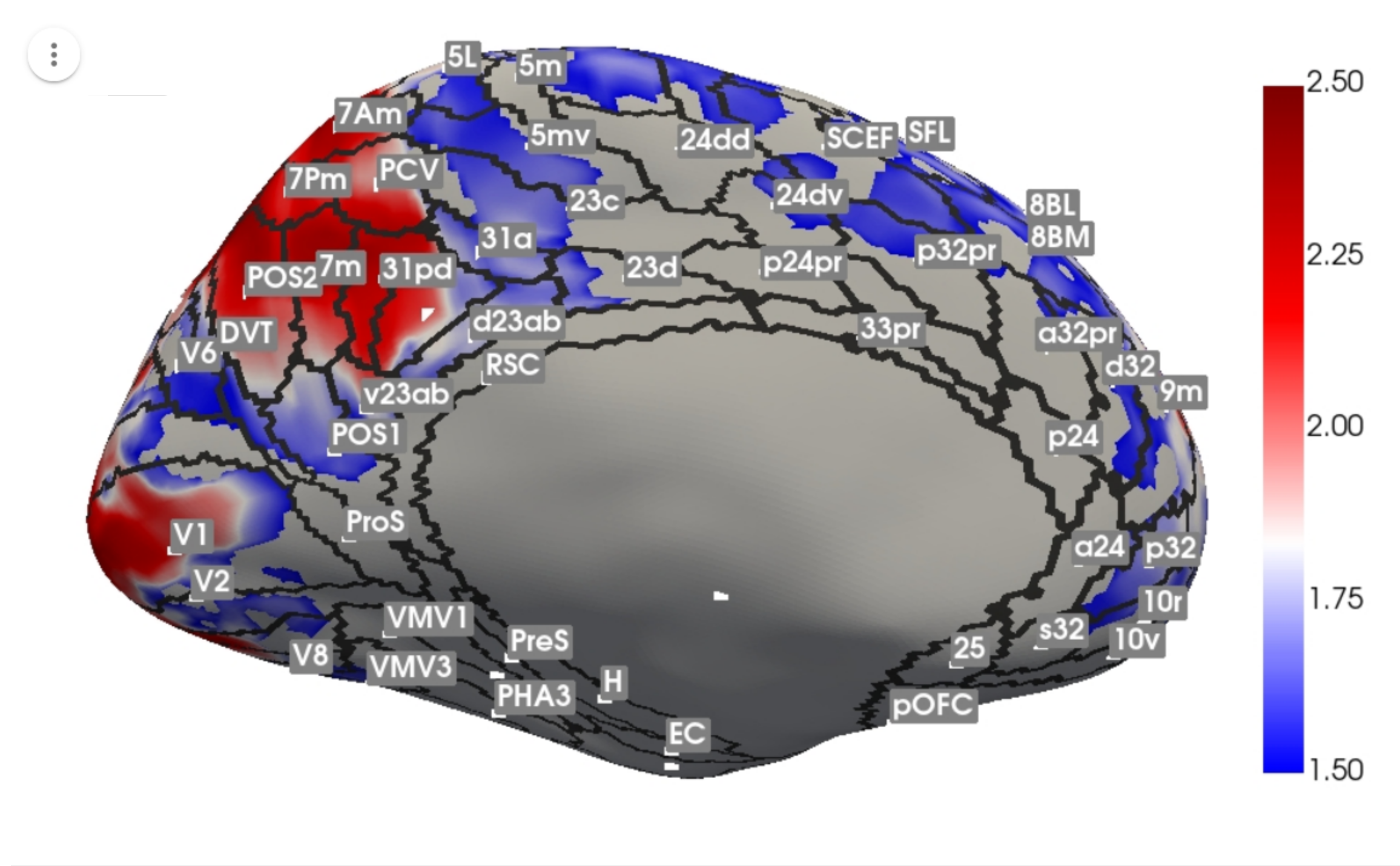}
\caption{Time constant map calculated directly from fMRI dataset with auto-correlation, the map is thresholded at 1.5s, which is the TR rate. The unit of the color map is second.}
\label{fig:temporal}
\end{figure}

In our analysis, we determined the cortex-wide autocorrelation time constant for the Narratives dataset using the approach outlined in Sec. \ref{sec:temporal}. Given that the time constant typically spans several seconds, we employed a maximum shift of 10 TRs to estimate the time constant, i.e., a total duration of 15 seconds. The resultant time constant map is illustrated in Fig. \ref{fig:temporal}. Upon examination, it's evident that distinct brain regions exhibit varying time constants. For areas unrelated to language processing, time constants were smaller. Beyond this, the time constants display a gradient, ascending from low to high, in alignment with the language hierarchy.

\subsection{Comparison of Hierarchical Rank}
\label{sec:rank}

While a visual comparison between Fig. \ref{fig:causal_diff} and Fig. \ref{fig:temporal} provides an intuitive sense of the hierarchical similarity, a numerical method to quantify this resemblance is preferred. We employ Spearman's rank correlation as a metric to gauge the similarity in hierarchical ranking.

Firstly, we need to designate the Regions of Interest (ROIs) for inclusion in our rank analysis. We delineate an ROI as a cerebral region within the Glasser atlas, that is effectively predicted by the language model. We establish a threshold, such that regions with a mean predictive accuracy exceeding this limit are incorporated. In ensuring that all pertinent cerebral zones are encompassed while excluding language-unrelated zones, we set our threshold at 0.06. This criterion results in the inclusion of 44 brain regions, constituting approximately one-fourth of the total regions in the Glasser atlas.

\begin{figure}[t]
\includegraphics[width=1.0\linewidth]{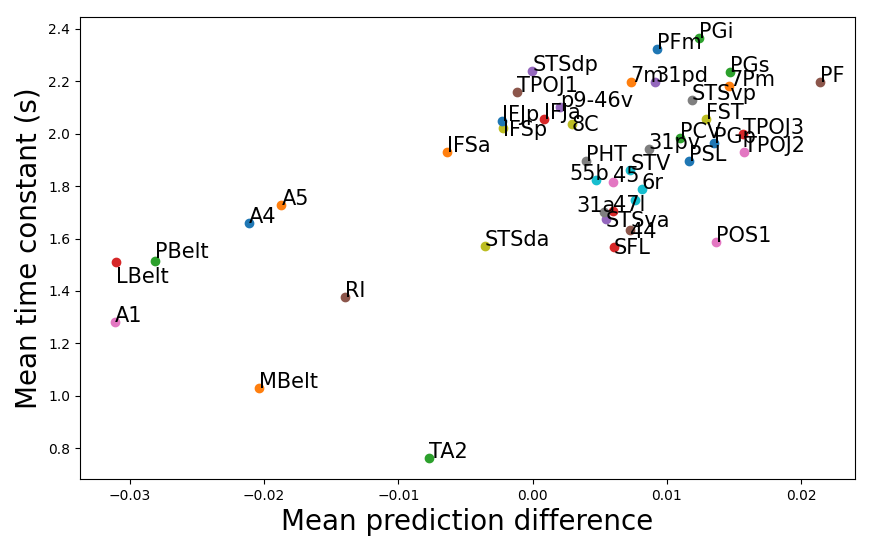}
\caption{Information integration index v.s. mean time constant per area. The information integration index is quantified by the difference in the mean prediction accuracy by high and low in-degree features of the language model.}
\label{fig:corrplot}
\end{figure}

Subsequent to this, we computed the information integration index of each brain region by the mean brain prediction accuracy difference based on Fig. \ref{fig:causal_diff}. Similarly, from Fig. \ref{fig:temporal}, we compute the average autocorrelation time constant specific to each region. Given our hypothesis that high in-degree features would be better at predicting regions higher in the hierarchy—regions expected to manifest longer time constants—we anticipate a positive correlation between the mean brain prediction accuracy difference and the mean time constant across our selected regions. The result is shown in \ref{fig:corrplot}. Aligning with our expectations, the resultant Spearman's rank correlation is 0.54, with a highly significant p-value of 0.00014.

\subsection{Time Constant of Language Features}

Our previous analyses highlight that the hierarchy in language model features, found out through cross-layer causality, relates to the hierarchy seen in brain signals during language tasks, as characterized by activity time scales. This suggests an intriguing interplay between information integration and temporal hierarchies. Given these findings, one would anticipate that low in-degree features in the language model would exhibit shorter activity time constants compared to high in-degree features.

% \begin{figure}[t]
% \includegraphics[width=1.0\linewidth]{LanguageFeatureT.png}
% \caption{Time constant for language features for high in-degree group in red and low in-degree group in blue.}
% \label{fig:langfeatureT}
% \end{figure}

\begin{figure}[t]
\includegraphics[width=1.0\linewidth]{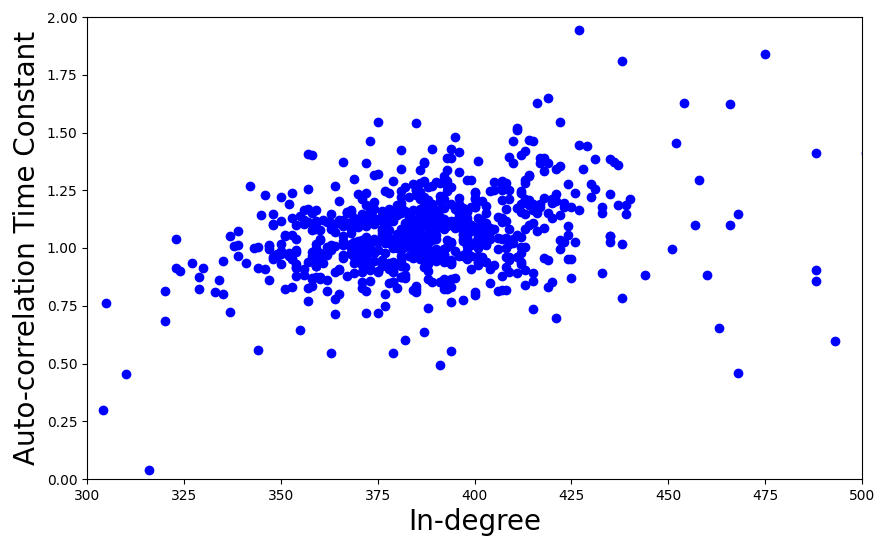}
\caption{In-degree v.s. auto-correlation time constant (unit: token) in the language model. The maximum lag is 50 tokens.}
\label{fig:langfeatureT}
\end{figure}

This expectation is verified in Fig. \ref{fig:langfeatureT}. Using the methodology described in Sec. \ref{sec:temporal}, we computed the activity time constants for each dimension of language features without applying PCA. The maximum lag we picked was 50 tokens, which roughly corresponds to 10 TRs. We then plot a figure in-degree as the horizontal axis and auto-correlation time constant as the vertical axis. 98\% of points have an in-degree within 300 to 500, and auto-correlation time constant below 2 tokens. It can be seen that in-degree is positively correlated with auto-correlation time constant. The resultant Spearman's rank correlation is 0.30, with a significant p-value of 1e-17.

In addition to the main results, we have included supplementary results in the Appendix. Sec. \ref{asec:layers} describes the reproduction of hierarchical maps using different layers. Section \ref{asec:sortwtime} presents a sanity check through hierarchical maps generated by grouping features based on time constants. Lastly, Section \ref{asec:lowfeature} demonstrates the creation of hierarchical maps using lower-layer features based on 'out-degree'.

\section{Discussion}

\begin{figure}[t]
\includegraphics[width=1.0\linewidth]{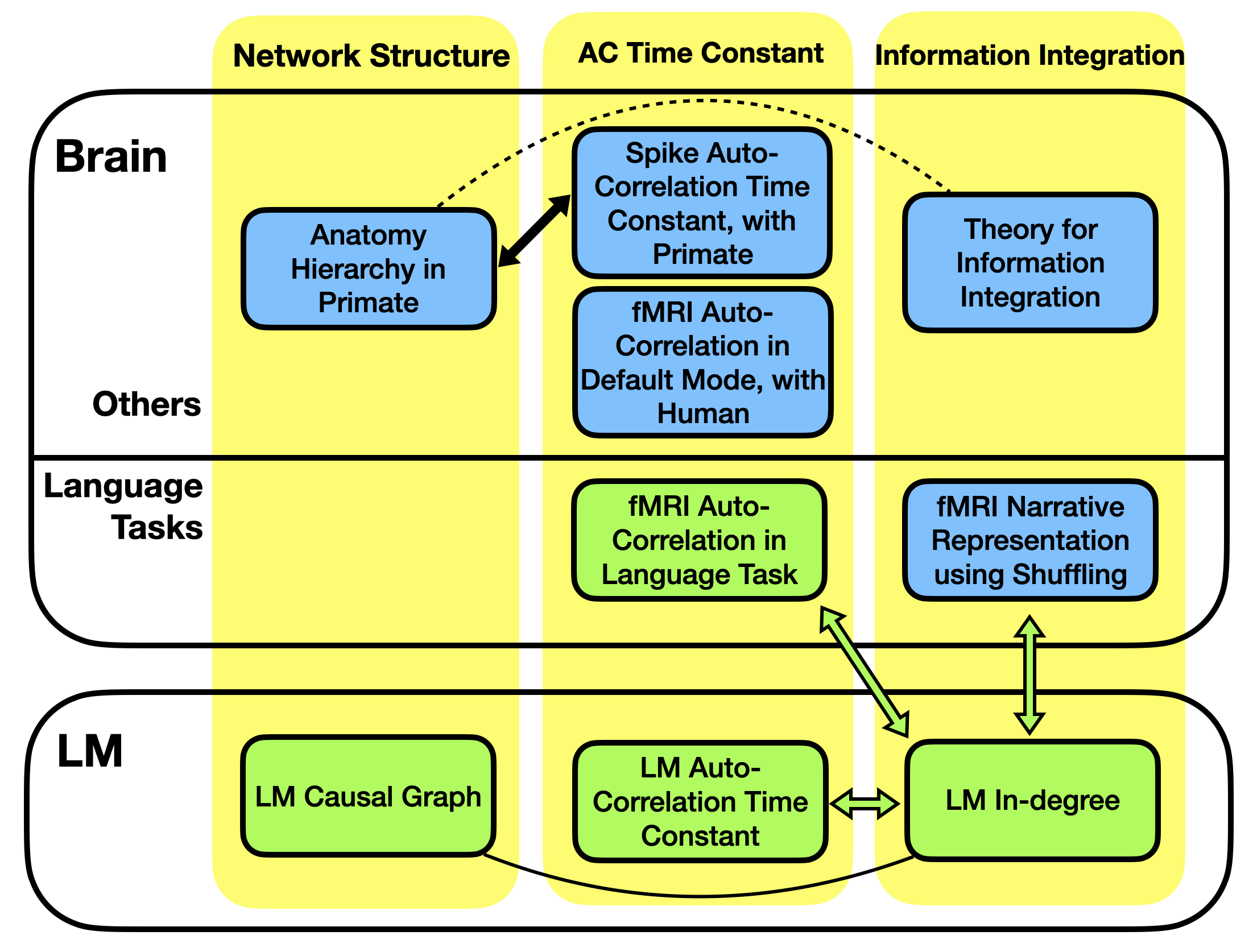}
\caption{Different concepts of hierarchy in part of previous works (shown in blue) and our work (shown in green).}
\label{fig:concepts}
\end{figure}

A central concept discussed in this paper is hierarchy. The notion of hierarchy, upon closer examination, reveals itself to be a concept of varied forms and interpretations. Figure \ref{fig:concepts} maps out various forms of hierarchies and their relations, incorporating elements from both prior studies and our current research. It is divided into two main sections: the upper section depicts hierarchies derived from brain studies, while the lower section focuses on those derived from language models. Hierarchies originating from the brain are further categorized based on their relevance to language tasks. Vertically, the figure illustrates three distinct hierarchy forms: network structure hierarchy, hierarchy inferred from auto-correlation-based time constants, and hierarchy based on information integration. The concepts from previous research are highlighted in blue boxes, whereas the green boxes denote the concepts introduced in our study.

The anatomical hierarchy, considered a 'gold standard' in hierarchical studies, is primarily observed in non-human primates. It has been shown to correlate with the spike auto-correlation time constant in the primate cortex, as indicated by the black arrow in our diagrams \citep{murray2014hierarchy}. Additionally, hierarchical gradients in the auto-correlation of resting-state fMRI signals have been identified \citep{raut2020hierarchical}. Theoretical frameworks, such as the workspace framework \citep{dehaene2001towards}, propose connections between information integration and anatomical hierarchy, represented in our figures by the dashed line. Building on this concept, there have been studies demonstrating the emergence of a cortical hierarchical map in language tasks. These studies utilized language patterns shuffled at different levels, highlighting the link between the hierarchical map in language tasks and the time window of information integration \citep{lerner2011topographic}.

In our study, we introduced corresponding new blocks of a language model into the figure. Firstly, we map the network structure of the brain onto the concept of a causal graph, and developed a measurement of information integration based on the in-degree of this causal graph. We demonstrated the ability of this measurement to reveal hierarchical structures captured through language shuffling techniques of previous work. Secondly, we found a correlation between the hierarchical map, as derived from the causal graph in-degree, and the fMRI auto-correlation time constant in language tasks. Lastly, our analysis of language model auto-correlation time constants revealed a correlation with the degree of information integration measured by in-degree. Collectively, these findings underscore a robust functional parallelism between language models and the human brain.

It is important to note that our figure does not encompass all relevant prior research. For instance, the study by Caucheteux et al. (2023) \citep{caucheteux2023evidence} rediscovers cortical hierarchy through predictive time windows, which is not included in our current representation.

\section{Conclusion}

This paper explores the relationship between the brain's hierarchy, intrinsic temporal scale, and information integration in the context of human natural language processing. Building upon the prevailing hypothesis that higher cortical areas typically exhibit longer time scales, and inspired by emerging research utilizing language models to study brain activity during language perception, we delve into the role of information integration in shaping hierarchy. By categorizing language features into low in-degree and high in-degree groups based on cross-layer causality, and subsequently using each group to predict brain activity, a distinct hierarchy based on the language model is observed. Specifically, low in-degree features correlate more with lower cortical areas, while high in-degree features align more with higher cortical areas. Intriguingly, the language model's auto-correlation time constant is also correlated with these features' in-degree, which is parallel to the gradient of the brain activity time constant. These findings suggest that the mapping between language model features and brain activity stems from similarity in information integration patterns rather than mere coincidental alignments.

%\section*{Acknowledgements}

\bibliography{narrative}
\bibliographystyle{acl_natbib}

\appendix

\section{Appendix}

\subsection{Robustness}
\label{asec:robustness}

In this section, we aim to demonstrate the robustness of our results across various model layers, scales, and types. 

\begin{figure}[t]
\includegraphics[width=1.0\linewidth]{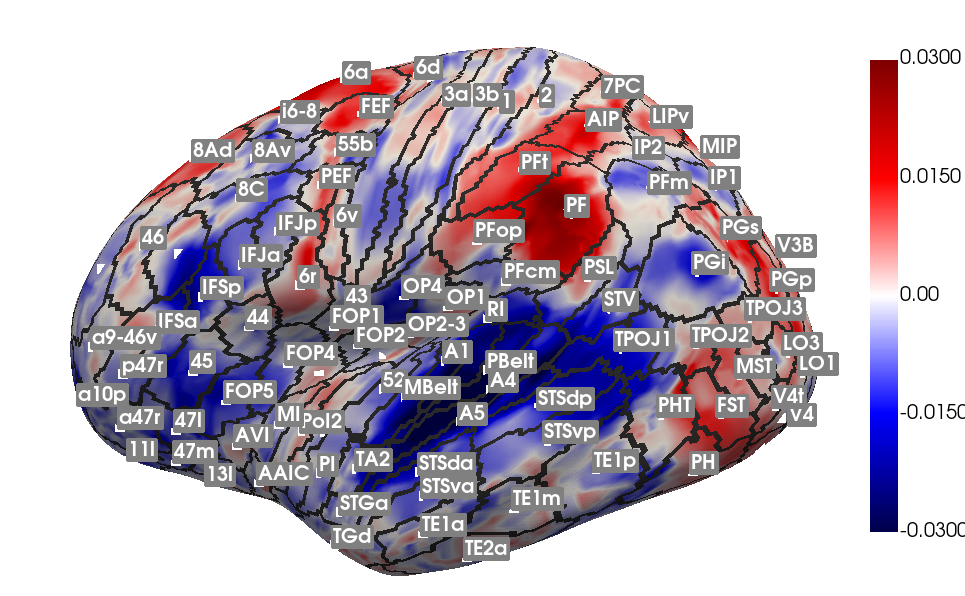}
\includegraphics[width=1.0\linewidth]{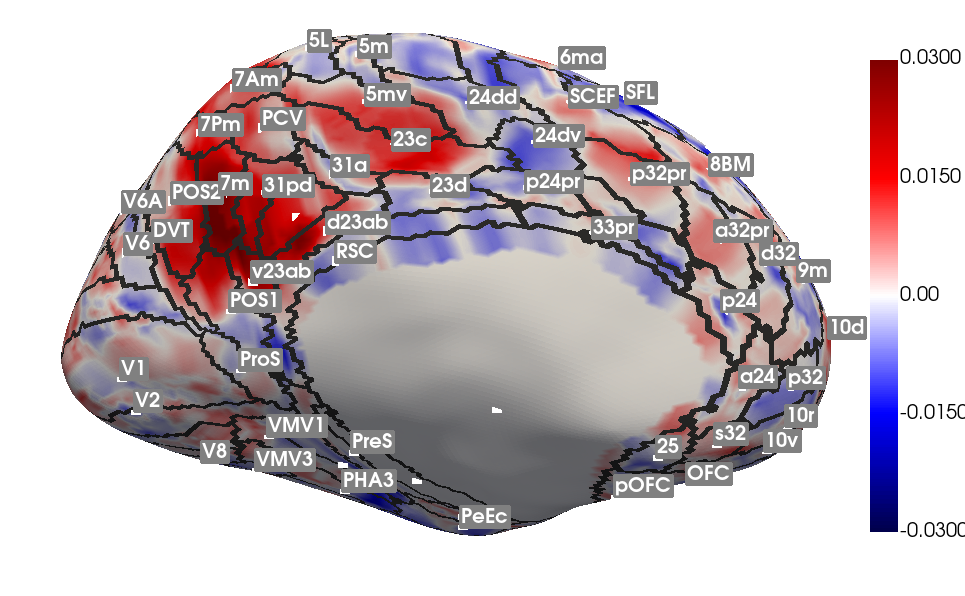}
\caption{Brain prediction accuracy difference between high in-degree feature group and low in-degree feature group measured with correlation, layer 8 of Opt-125m..}
\label{fig:causal_diff_l7}
\end{figure}

\begin{figure}[t]
\includegraphics[width=1.0\linewidth]{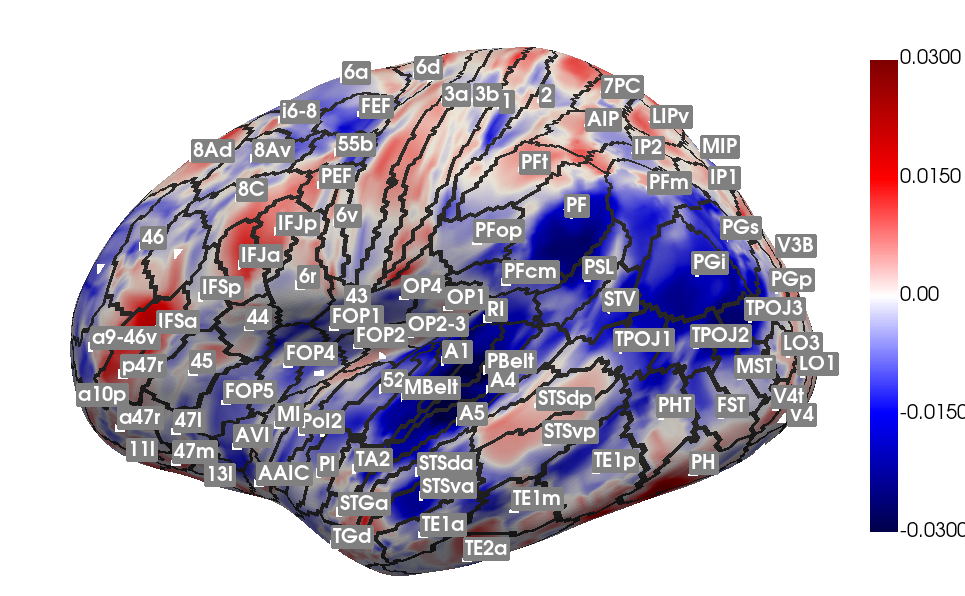}
\includegraphics[width=1.0\linewidth]{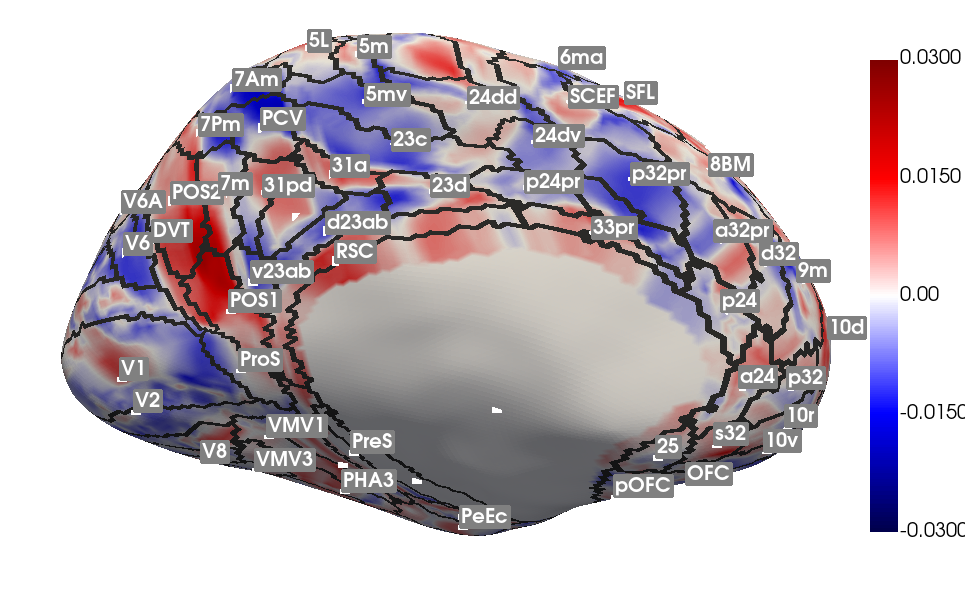}
\caption{Brain prediction accuracy difference between high in-degree feature group and low in-degree feature group measured with correlation, layer 12 of Opt-350m.}
\label{fig:causal_diff_350m}
\end{figure}

\begin{figure}[t]
\includegraphics[width=1.0\linewidth]{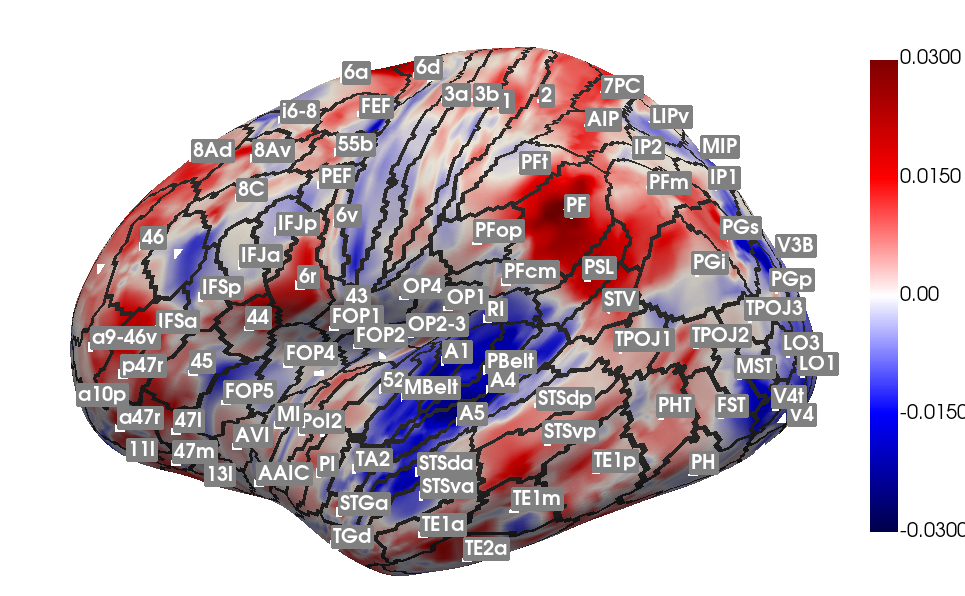}
\includegraphics[width=1.0\linewidth]{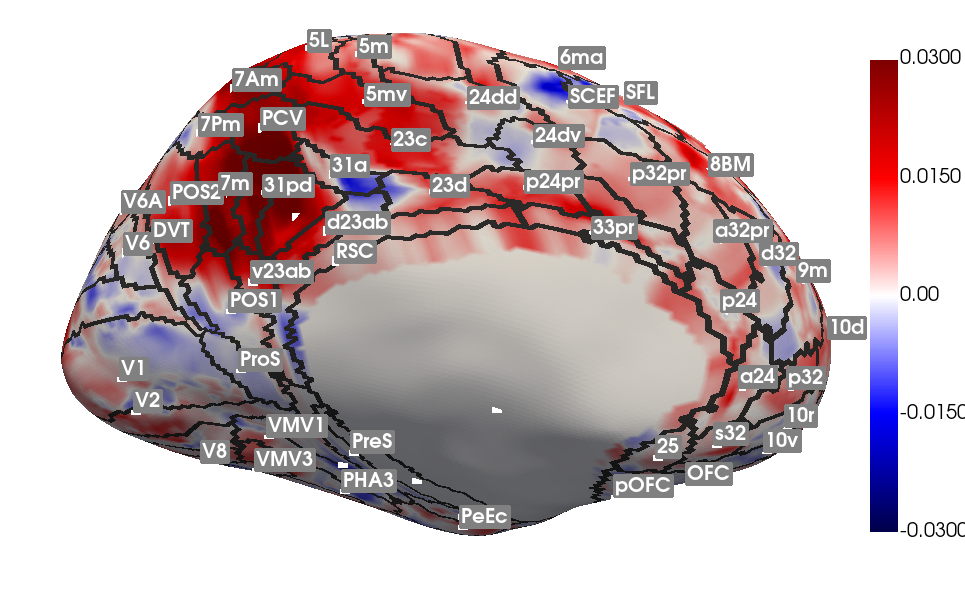}
\caption{Brain prediction accuracy difference between high in-degree feature group and low in-degree feature group measured with correlation, layer 9 of Gpt2.}
\label{fig:causal_diff_gpt2}
\end{figure}

In our primary manuscript, we utilized layer 9 of Opt-125m due to its superior brain prediction capabilities. To validate the consistency of our findings, we also examine neighboring layers, such as layer 8. Figure \ref{fig:causal_diff_l7} depicts the difference in brain prediction accuracy between the high in-degree and low in-degree feature groups of layer 8 when paired with layer 3. Furthermore, we assessed our results using the larger Opt-350m language model, as illustrated in Fig. \ref{fig:causal_diff_350m} for layers 6 and 12. We also applied our methodology to GPT2, examining layers 4 and 9, with the outcomes presented in Fig. \ref{fig:causal_diff_gpt2}. The hierarchical rank was determined using the approach detailed in Sec. \ref{sec:rank}, and the collective results are summarized in Table \ref{tab:robust}.

\begin{table*}[ht]
\caption{Spearman correlation of calculated hierarchical rank among different models. All entries except for those inside parenthesis has p-value smaller than 0.05.}
\label{tab:robust}
% \vskip 0.15in
\begin{center}
\begin{small}
%\begin{sc}
%\resizebox{\textwidth}{!}{
\begin{tabular}{l|cccccc}\hline
Feature       & Opt125m L8 &   Opt125m L9     & Opt350m L12    & GPT2 L9   & Time Constant\\
\hline
%Opt125m L8     & 1.0       &   0.87           &        0.60    &  0.80  & 0.50 \\
Opt125m L8     & 1.0       &   0.81           &        0.31    &  0.58  & 0.36 \\
%Opt125m L9     & 0.87      &   1.0            &        0.54    &  0.73  & 0.50 \\
Opt125m L9     & 0.81      &   1.0            &        (0.08)    &  0.66  & 0.54 \\
%Opt350m L12    & 0.60      &   0.54           &        1.0     &  0.56  & 0.33 \\
Opt350m L12    & 0.31      &   (0.08)           &        1.0     &  0.39  & (0.23) \\
%GPT2 L9        & 0.80      &   0.73           &        0.56    &  1.0   & 0.59 \\
GPT2 L9        & 0.58      &   0.66           &        0.39    &  1.0   & 0.47 \\
%Time Constant  & 0.50      &   0.50           &        0.33    &  0.59  & 1.0  \\
Time Constant  & 0.36      &   0.54           &        (0.23)    &  0.47  & 1.0  \\

\hline
\end{tabular}
%}
%\end{sc}
\end{small}
\end{center}
% \vskip -0.1in
\end{table*}

\subsection{Hierarchy from Layers}
\label{asec:layers}

Besides information integration and activity time constants, existing literature has discussed the relationship between brain hierarchy and the layers of language models \citep{toneva2019interpreting, goldstein2022correspondence}. We further validated these findings using the Narratives dataset.

\begin{figure}[t]
\includegraphics[width=1.0\linewidth]{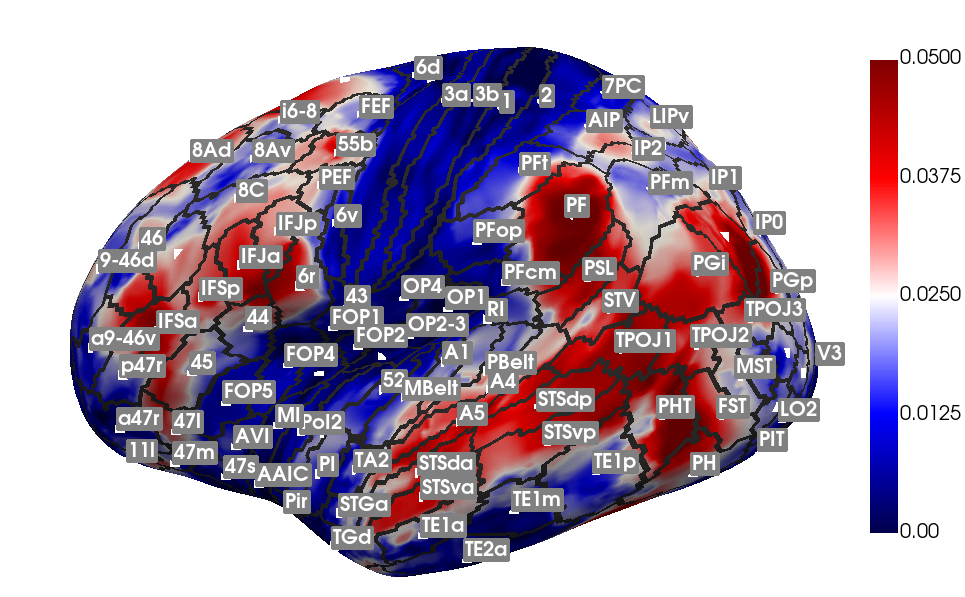}
\includegraphics[width=1.0\linewidth]{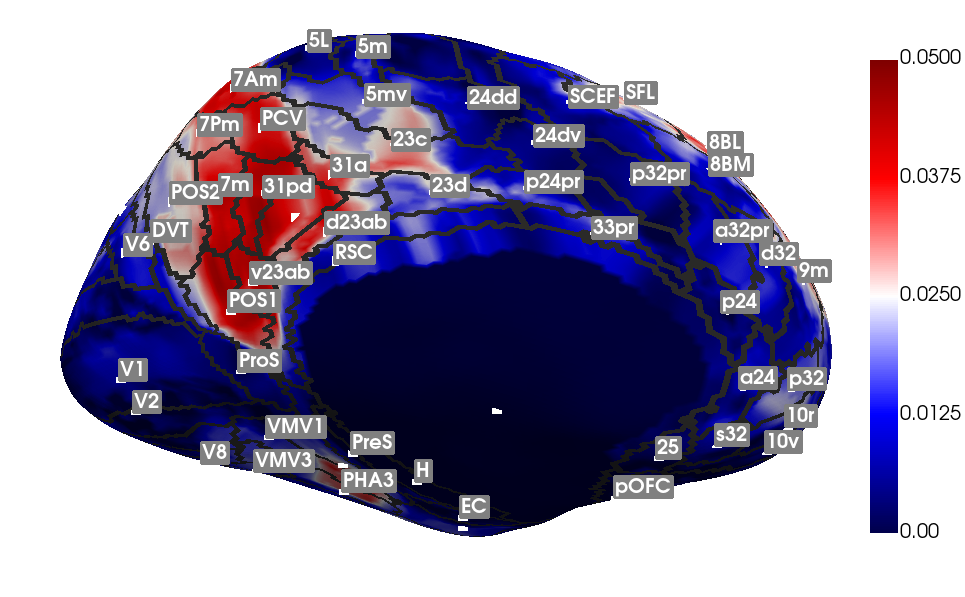}
\caption{Brain hierarchy calculated by precision map subtraction between layer 9 and layer 1 of Opt 125m language model features.}
\label{fig:layers}
\end{figure}

Fig. \ref{fig:layers} displays the differential map of brain prediction accuracy between layers 9 and 1 of the OPT-125m language model features (specifically, the result of subtracting layer 1 from layer 9). This pattern aligns with those derived from causality and activity time constants. However, differences are evident. As the layer number ascends, there is minimal decline in prediction accuracy for lower hierarchy areas. Conversely, there's a significant surge in accuracy for higher hierarchy regions, leading to a nearly monotonic increase in prediction accuracy from layer 1 to layer 9. This suggests that the representations in layer 9 contains information to predict both lower and higher hierarchy brain regions.

\begin{figure}[t]
\includegraphics[width=1.0\linewidth]{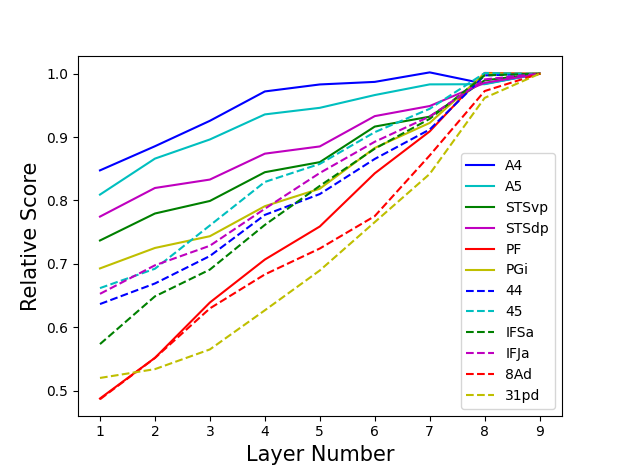}
\caption{Normalized average brain prediction accuracy for each ROI region along number of layers.}
\label{fig:layer_scale}
\end{figure}

To substantiate this hypothesis, we plotted the ROI average brain prediction precision across the layers of OPT 125m, spanning from layer 1 to layer 9, as illustrated in Fig. \ref{fig:layer_scale}. The plotted accuracy is normalized by taking its ratio to the accuracy of layer 9 of each region. The results indicate that lower-order regions, such as A4 and A5, are effectively predicted by layer 1, whereas higher cortical areas, like PF and 31pd, benefit more from higher layers.

%\subsection{Language Network Corresponds to Low Level}

\subsection{Sorting language features with time constant}
\label{asec:sortwtime}

Our main result reported in Sec. \ref{sec:causal_res} groups features based on information integration. And we related the calculated hierarchy with that calculated from activity time scale. As a straight forward sanity check, if we group features also based on activity time constant of language features, brain hierarchy would also expected to emerge.

\begin{figure}[t]
\includegraphics[width=1.0\linewidth]{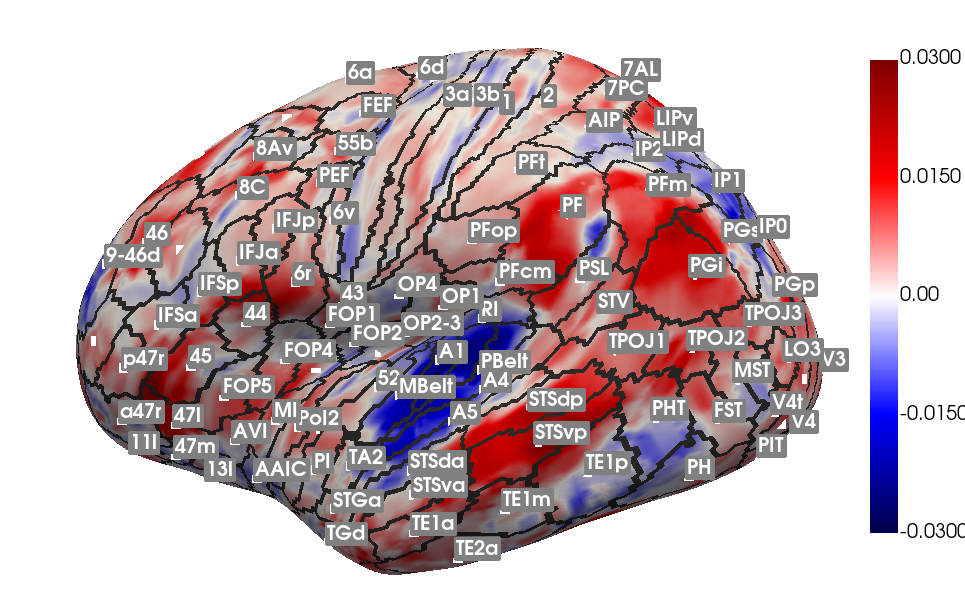}
\includegraphics[width=1.0\linewidth]{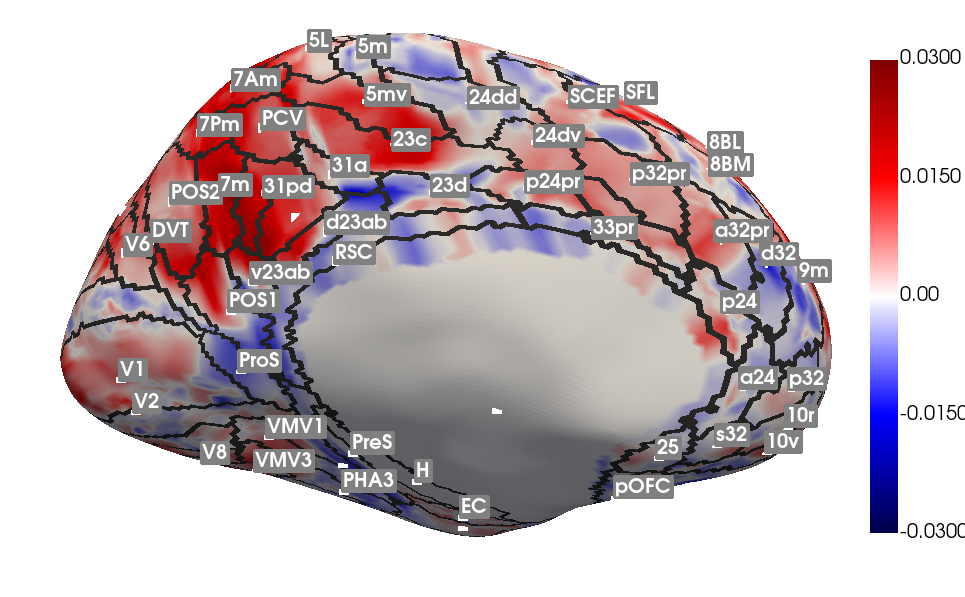}
\caption{Brain hierarchy calculated by precision map subtraction between slow group and fast group (slow minus fast) of Opt-125m language model features.}
\label{fig:time_diff}
\end{figure}

The result is shown in Fig. \ref{fig:time_diff}. It can be seen that brain prediction accuracy difference from feature group with different time scale can also capture cortical hierarchy. Where fast features predict lower cortical regions like A4, A5 better, while slow features predict higher cortical regions like PF, 31pd better.

Our principal findings presented in Sec. \ref{sec:causal_res} categorize features based on causality. We then correlated the derived hierarchy with that calculated from the activity time scale. As a validation, one would anticipate that by grouping features based on the activity time constant of language features, the brain hierarchy would also become evident.

This observation is illustrated in Fig. \ref{fig:time_diff}. The difference in brain prediction accuracy among feature groups with varying time scales delineates the cortical hierarchy. Specifically, features with faster time scales more accurately predict lower-order regions such as A4 and A5, whereas features characterized by slower time scales are better suited to predicting higher cortical areas like PF and 31pd.

\subsection{Features from Low Layers}
\label{asec:lowfeature}

\begin{figure}[t]
\includegraphics[width=1.0\linewidth]{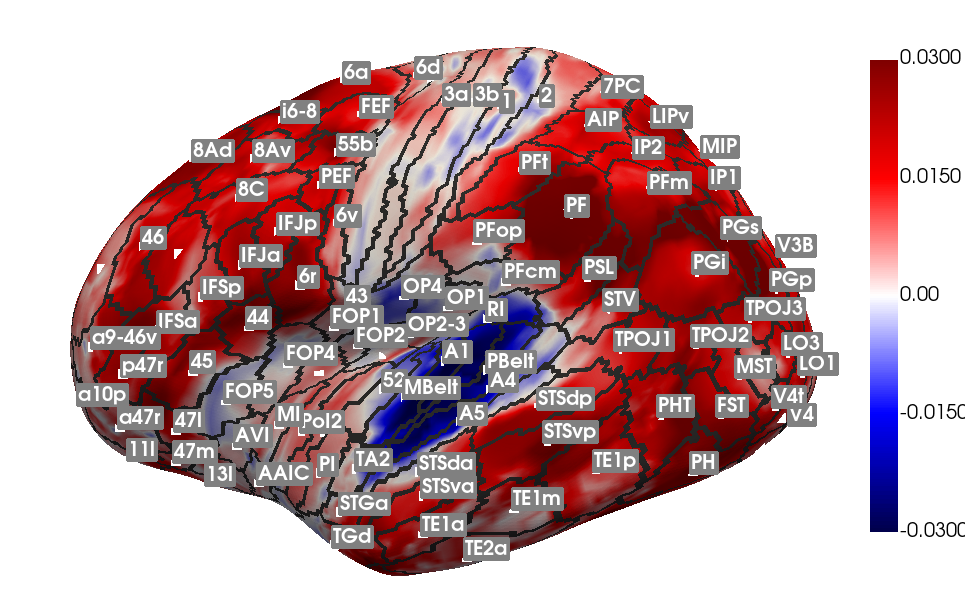}
\includegraphics[width=1.0\linewidth]{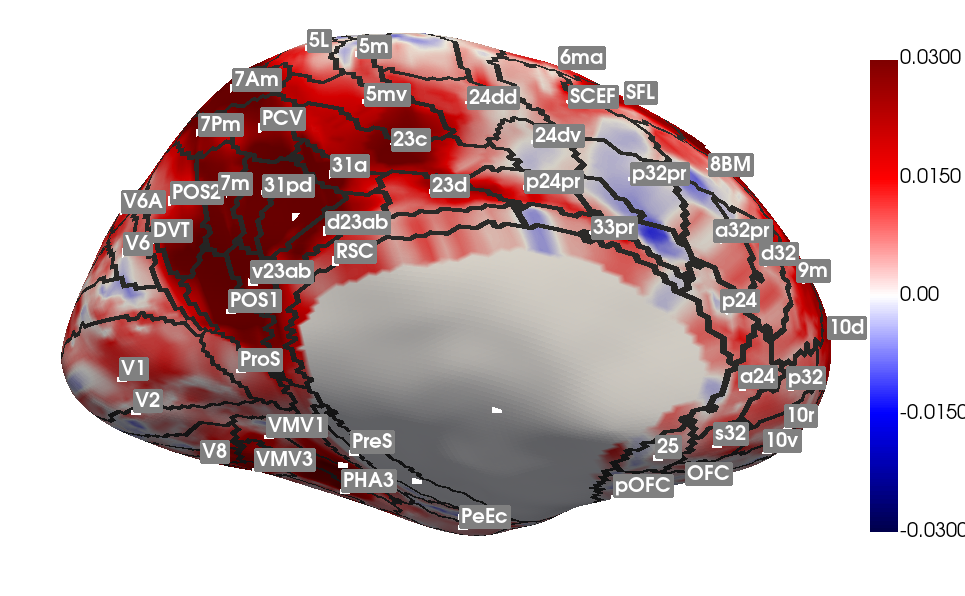}
\caption{Brain prediction accuracy difference between "high in-degrees" feature group of layer 9 and "high out-degrees" feature group of layer 4, measured with correlation.}
\label{fig:causal_diff_multil}
\end{figure}

In the main manuscript, we segregated the features of layer 9 based on in-degree measures using a causal graph measure determined between layers 4 and 9. This demonstrated that the delineated features align with the cortical hierarchy during language processing. Our preference for layer 9 stems from its optimal fit with brain data. Notably, the causality matrix can also be applied to partition the features of layer 4 using "out-degree". Here, features are partitioned into groups with "low out-degrees" and "high out-degrees". We expected that features with "high out-degree" may excel at predicting low cortical area. As described in Sec. \ref{asec:layers}, earlier layers adequately predict activity in lower cortical regions. To validate our approach, we expect that if we utilize "high out-degrees" features from layer 4 combined with "high in-degrees" features from layer 9, the cortical hierarchy would also emerge. As depicted in Fig. \ref{fig:causal_diff_multil}, our primary conclusion remains valid. The calculated Spearman's rank correlation with the time constant map is 0.57, p-value is 5e-5.

We also present results derived exclusively from features of layer 4. While this layer hasn't fully developed features that adeptly predict brain activity, especially in higher cortical areas, examining the correlation map differences between its "low out-degree" and "high out-degree" features remains insightful. The findings are illustrated in Fig. \ref{fig:causal_diff_multil}. The associated Spearman's rank correlation with the time constant map stands at 0.37, p-value is 0.015.

\begin{figure}[t]
\includegraphics[width=1.0\linewidth]{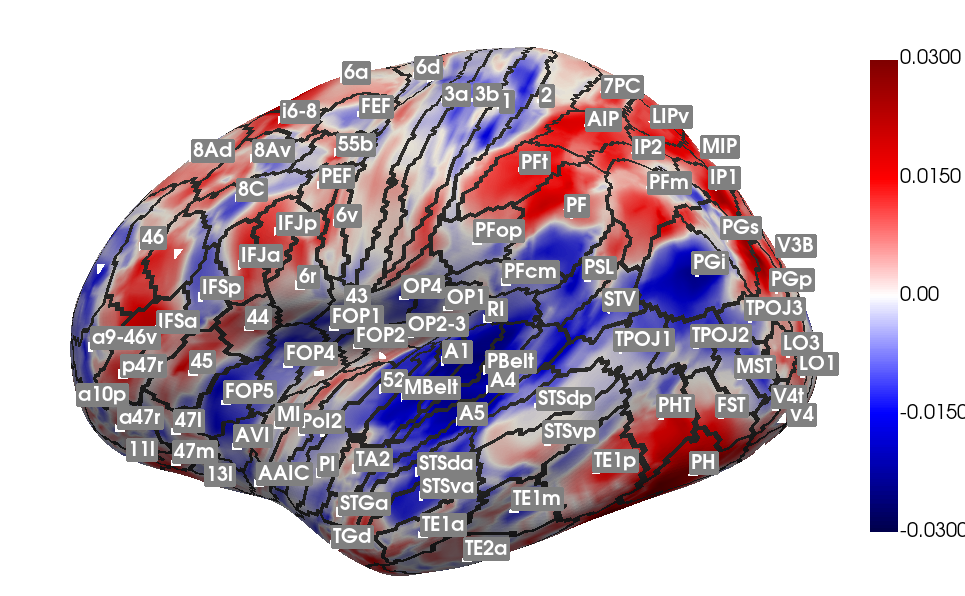}
\includegraphics[width=1.0\linewidth]{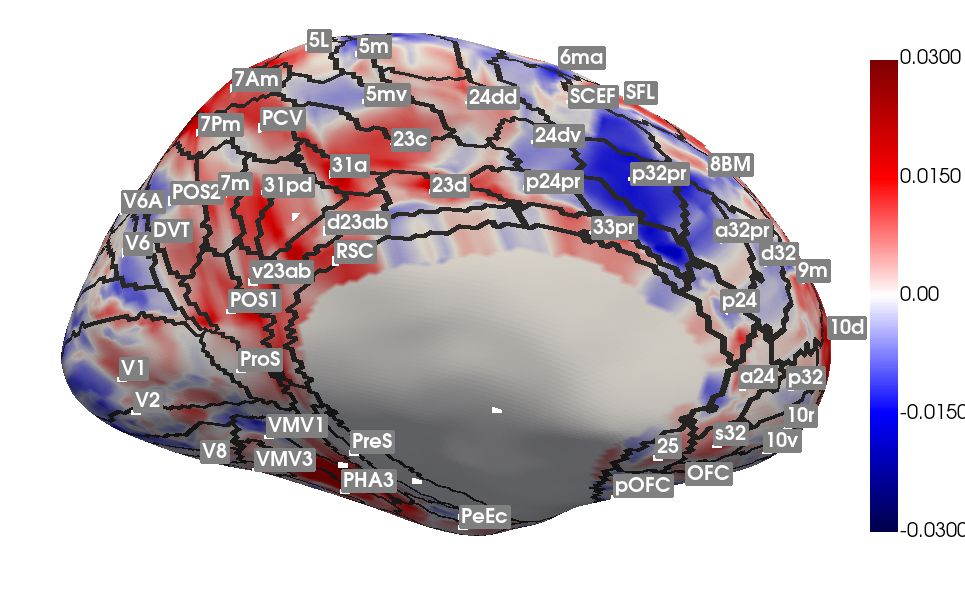}
\caption{Brain prediction accuracy difference between "low out-degrees" feature group and "high out-degrees" feature group of layer 4, measured with correlation.}
\label{fig:causal_diff_l4}
\end{figure}

\end{document}